%% file: main.tex
\documentclass[sigconf]{acmart}

\AtBeginDocument{%
  }

\setcopyright{acmlicensed}
\copyrightyear{2018}
\acmYear{2018}
\acmDOI{XXXXXXX.XXXXXXX}

\usepackage{algorithm}
\usepackage{algorithmic}

\begin{document}

\title{Adaptive Conditional Expert Selection Network for Multi-domain Recommendation}

\author{Kuiyao Dong}
\authornote{Both authors contributed equally to this research.}
\email{dongkuiyao@oppo.com}
\affiliation{%
  \institution{OPPO}
  \streetaddress{}
  \city{Shenzhen}
  \state{Guangdong}
  \country{China}
  \postcode{}
}

\author{Xingyu Lou}
\authornotemark[1]
\email{louxingyu@oppo.com}
\affiliation{%
  \institution{OPPO}
  \streetaddress{}
  \city{Shenzhen}
  \state{Guangdong}
  \country{China}
  \postcode{}
}

\author{Feng Liu}
\email{liufeng4@oppo.com}
\affiliation{%
  \institution{OPPO}
  \streetaddress{}
  \city{Shenzhen}
  \state{Guangdong}
  \country{China}
  \postcode{}
}

\author{Ruian Wang}
\email{wangruian@oppo.com}
\affiliation{%
  \institution{OPPO}
  \streetaddress{}
  \city{Shenzhen}
  \state{Guangdong}
  \country{China}
  \postcode{}
}

\author{Wenyi Yu}
\email{yuwenyi@oppo.com}
\affiliation{%
  \institution{OPPO}
  \streetaddress{}
  \city{Shenzhen}
  \state{Guangdong}
  \country{China}
  \postcode{}
}

\author{Ping Wang}
\email{ping.wang@oppo.com}
\affiliation{%
  \institution{OPPO}
  \streetaddress{}
  \city{Shenzhen}
  \state{Guangdong}
  \country{China}
  \postcode{}
}

\author{Jun Wang}
\email{junwang.lu@gmail.com}
\authornote{Corresponding author.}
\affiliation{%
  \institution{OPPO}
  \streetaddress{}
  \city{Shenzhen}
  \state{Guangdong}
  \country{China}
  \postcode{}
}

\renewcommand{\shortauthors}{Dong et al.}


\begin{abstract}
Mixture-of-Experts (MOE) has recently become the de facto standard in Multi-domain recommendation (MDR) due to its powerful expressive ability. However, such MOE-based method typically employs all experts for each instance, leading to scalability issue and low-discriminability between domains and experts. Furthermore, the design of commonly used domain-specific networks exacerbates the scalability issues.
To tackle the problems, We propose a novel method named \textbf{CESAA} consists of \textbf{C}onditional \textbf{E}xpert \textbf{S}election (CES) Module and \textbf{A}daptive Expert \textbf{A}ggregation (AEA) Module to tackle these challenges.
Specifically, CES first combines a sparse gating strategy with domain-shared experts. Then AEA utilizes mutual information loss to strengthen the correlations between experts and specific domains, and significantly improve the distinction between experts. 
As a result, only domain-shared experts and selected domain-specific experts are activated for each instance, striking a balance between computational efficiency and model performance. Experimental results on both public ranking and industrial retrieval datasets verify the effectiveness of our method in MDR tasks.
\end{abstract}




\maketitle

\input{1.introduction}

\input{2.method}

\input{3.experiments}

\section{CONCLUSION AND FUTURE WORKS}
This paper proposed an adaptive multi-domain modeling method, CESAA, which leverages sparse selection and mutual information constraints for multi-domain modeling. Extensive experiments conducted on both retrieval stage and ranking stage datasets 
have validated the effectiveness and efficiency of the CESAA model. 
Future works could extend sparse expert selection to the shared experts and the number of selected experts $\mathbf{k}$ could be personalized based on domain characteristics.

\bibliographystyle{ACM-Reference-Format}
\bibliography{references}

\end{document}

%% file: 1.introduction.tex
\section{INTRODUCTION}

Click-through rate (CTR) prediction is a critical application of modern recommendation systems. Traditional CTR models have primarily focused on single domains, resulting in under-utilization of data and high maintenance costs. 


The challenges of MDR models lie in efficiently and effectively utilizing data while capturing the heterogeneity and commonalities across different domains. When simply merging data from different domains, neglecting variations in data distribution and user behavior across contexts can lead to sub-optimal performance.
To address this issue, inspired by multi-task learning (MTL)\cite{ma2018mmoe}\cite{tang2020ple}\cite{xu2022mvke}, numerous multi tower-based approaches have emerged in recent years, which primarily utilize the Mixture-of-Experts (MoE) \cite{ma2018mmoe} structure 
to effectively leverage shared information and alleviate negative transfer issues
by combining shared bottom network with multiple domain-specific expert towers. Overall, The objective is to ensure that each expert tower learns scene-specific content in a discriminative manner.

Despite their effectiveness, previous MoE-based works\cite{jiang2022adin}\cite{li2022adaptdhm}\cite{tang2020ple} still face the following challenges.
The first is the \textbf{low discriminability} problem, which is common but non-trivial when employing multiple experts in MDR. Since each domain utilizes all experts, differing only in their gating weights, it struggles to distinguish between the shared characteristics and unique differences among the experts. Some domain-specific works like ADIN \cite{jiang2022adin} could improve experts' discriminability,  but they may encounter the second challenge: \textbf{Scalability}. In most previous MoE-based works, all experts are activated for each input sample. As the number of domains and experts increases, the time consumption for training and inference increases significantly.
More importantly, these approaches heavily depend on pre-defined domain partitions that are based on rules or expert knowledge. This limitation implies that the quality of prior experience significantly constrains the expressive capability of the models. For instance, in-feed ads are displayed on over 30 platforms of OPPO smartphones (e.g., browser app and weather app). It is characterized by the exhibition of ads in the form of large images, small images, image galleries, or videos within the browser.  These scenes exhibit different user behaviors and data distributions (e.g., daily exposure counts, user click-through rates, and exposure rates). Recently, some works like AdaptDHM\cite{li2022adaptdhm}
utilize the clustering method to avoid manual domain definition. However, it still faces the problem of low discriminability.
When modeling, we must consider the commonalities and heterogeneities between different scenes. However, the boundaries between scenes become increasingly blurred, and relying solely on manually defined domain indicators (e.g., ID) without considering the potential and complex correlations among domains will result in sub-optimal performance.

To overcome these challenges simultaneously, we propose \textbf{CESAA}, a novel and flexible method comprising two core modules: a Conditional Expert Selection (CES) Module and an Adaptive Expert Aggregation (AEA) Module.
Specifically, CES is designed to tackle differentiation and scalability problems by incorporating a sparsity constraint, which allows each expert to focus only on part of domains or features. Furthermore, the shared expert is shared by all domains and is responsible for learning the commonalities between domains.
To strengthen the correlation between experts and specific domains, AEA leverages mutual information loss to enhance the modularity and discriminative power of the experts. This means there is no need for manually pre-defined and fine-grained domain partitioning. Instead, the optimal partitioning can be discovered through end-to-end learning.
  

In summary, our paper makes the following contributions.
\begin{itemize}
\item We propose a model named CESAA with \emph{sparse selection} and \emph{adaptive aggregation} mechanisms to solve low-discriminability and scalability problems simultaneously in multi-domain recommendation.
\item We design an adaptive mutual information-guided expert aggregation module in CESAA. This mechanism facilitates the effective aggregation of similar domains through end-to-end learning. 
    
\item We perform extensive experiments on a self-built retrieval dataset and an open-source ranking dataset. The experimental results indicate that CESAA performs better than the state-of-the-art approaches.
\end{itemize}

%% file: 2.method.tex
\section{Methodology}
\subsection{Problem Formulation}
Given the label space $\mathcal{Y}$ and the feature space $\mathcal{X}$ consisting of domain indicator features and domain-agnostic features (e.g., user features, item features, and context features), multi-domain recommendation is to construct a unified CTR prediction function \(\mathcal{F}:\mathcal{X} \rightarrow \mathcal{Y}\) for a set of business domains \({\{D_m\}}_{m=1}^{M}\), which could reduce resource consumption and improve prediction accuracy. MoE architectures are widely applied with multiple expert networks: \(\{E\}_{i=1}^{N}\). This paper matches data from different domains to distinct expert groups with mutual information constraints by defining a gating function \(\mathcal{G}: \mathcal{X} \rightarrow KeepTopK(\{E\}_{i=1}^{N}) \) and a mutual information loss \(  \mathcal{L}^{MI}\) between domains and experts. This allows us to effectively capture each domain's characteristics and improve the model's overall performance.

\subsection{Architecture Overview}
\begin{figure}[!t]
  \centering
  \includegraphics[width=1\columnwidth]{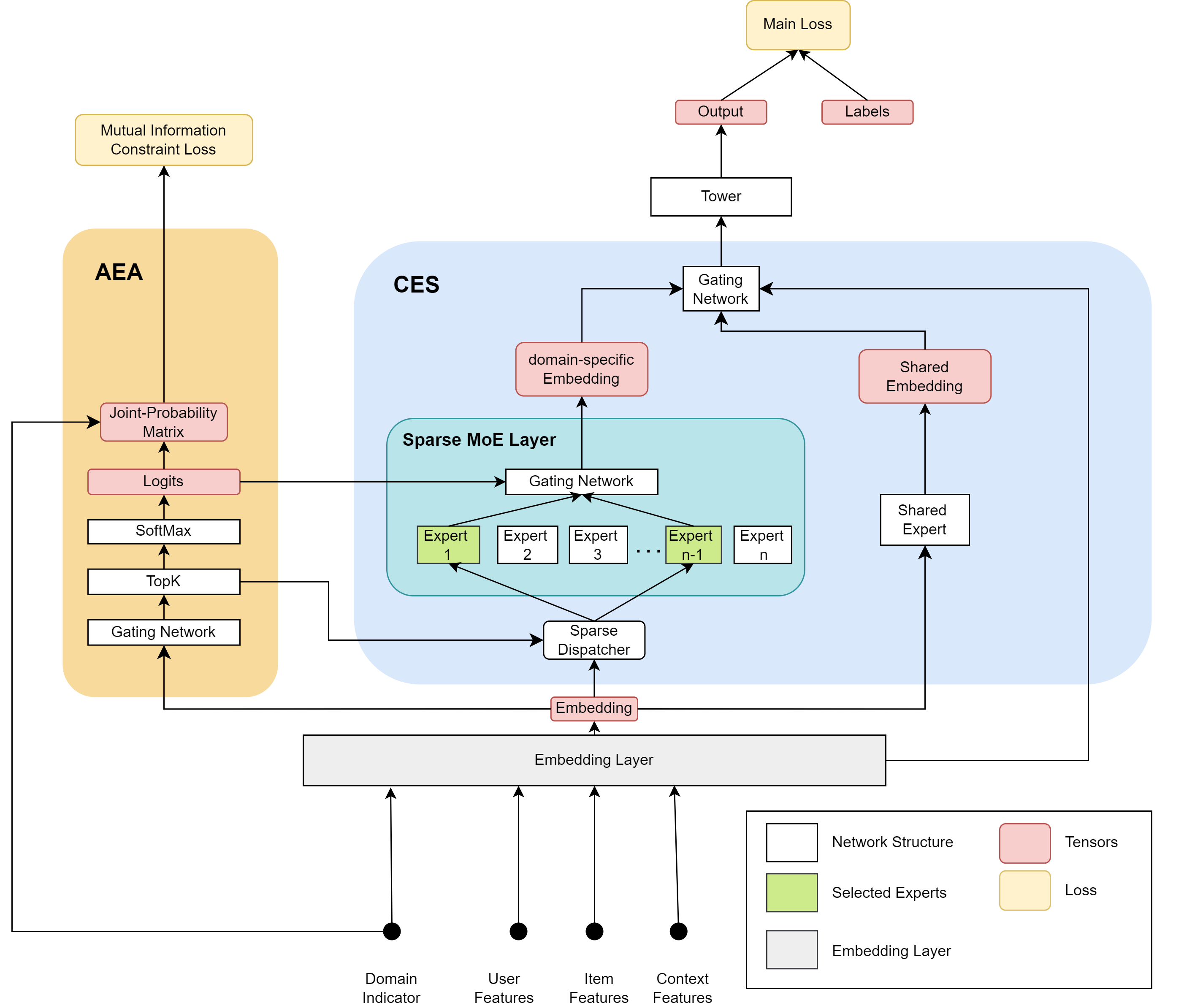}
  \caption{Overview of CESAA model architecture. CES consists of a sparse MoE and a shared expert. AEA utilizes a mutual information constraint loss to strengthen the dependency between experts and domains.}
  \label{fig:model_architecture}
\end{figure}

We present the details of our proposed model \textbf{CESAA}. As shown in Figure \ref{fig:model_architecture}, a batch of multi-domain data is fed into the embedding layer to be transformed into dense embeddings. Then \textbf{AEA} is designed to determine which group of specific experts will be chosen for the corresponding instances by maximizing mutual information between experts and domains. Finally, \textbf{CES} takes these embeddings as input and combines sparsely selected and shared experts to generate the predicted results.



\subsection{Conditional Expert Selection Module}
\label{sec:cag_cons&pref_mod}

In this section, we propose the CES, an innovative Sparse Mixture of Experts (MoE) strategy, which aims to improve the efficiency and differentiation of expert networks in multi-domain recommendation systems. 

\subsubsection{\textbf{Sparse MoE}} 

The Sparse Gating strategy activates only the top-k expert networks most relevant to the input, promoting sparsity and efficiency\cite{shazeer2017outrageously, zhou2022mixture}.  The following Eq. ( \ref{eq:smoe_1}-\ref{eq:smoe_3} )\cite{shazeer2017outrageously} outlines the structure of the Sparse MoE model.  Sparse MoE utilizes Noisy Top-K Gating, which applies adjustable Gaussian noise and retains only the top-k experts for each input of $x$. 





\begin{equation}\label{eq:smoe_1}
G(x) = Softmax(KeepTopK(H(x), k)) , 
\end{equation}

\begin{equation}\label{eq:smoe_2}
H(x)_i = (x \cdot W_g )_i + StandardNormal() \cdot Softplus((x \cdot W_{noise})_i) , 
\end{equation}

\begin{equation}\label{eq:smoe_3}
   KeepTopK(v, k)_i = 
    \begin{cases}
        v_i & \text{if $v_i$ in top k of $v$} \\
        -\infty & \text{otherwise}
    \end{cases} 
\end{equation}




\subsubsection{\textbf{ Shared Expert}}\label{sec:sgc}
Even though Sparse MoE \(E_{sp}\) could reduce the computing resources while the number of experts scales up, we argue that it may lose the common information across different domains. Inspired by \cite{tang2020ple}, we additionally introduce a shared expert layer \(E_{sh}\), so as to capture the commonalities between experts.  \(G(x)\) denotes the gating network to fuse the outputs of sparse MoE and share experts. Finally, the output of the CES layer is formulated as follows:

\begin{equation}\label{eq:scgc}
\hat{y}(x) = Sigmoid(G(x)\cdot ( E_{sp}(x) \mathbin\Vert E_{sh}(x))) , 
\end{equation}

\begin{equation}
    \mathcal{L}^{BCE} = -\frac{1}{N}\sum_{i=1}^N y_{i} \log(\hat{y}_{i})+ (1- y_{i}) \log(1-\hat{y}_{i})) . 
\end{equation}
where \(\hat{y}_{i}\) is the output of CESAA, \(y_{i}\) is the label of i- th sample. \(y_{i}=1\)  indicates the current user will click the recommended item and 0 otherwise.


\subsection{Adaptive Expert Aggregation Module }\label{sec:mutial_information_loss}

We incorporate mutual information loss into the training procedure to further strengthen the association between experts and domains. This loss function encourages each expert to specialize in a specific domain by increasing the mutual information between its output and the target domain labels \cite{chen2023mod}. 

For a given MoE layer, \(P(E_i | D_j)\) is defined as the frequency with which expert \(E_i\) is assigned to samples in Domain \(D_j\) by the routing network. 
For example, in domain  \(D_j\), if there are 100 samples and the routing network assigns 40 of them to expert \(E_i\), then \(P(E_i | D_j)\) = 0.4. The network assigns soft weights to experts instead of hard assignments, so we can add these weights to measure expert frequency. 

One important insight in our research is that experts should have some domain specialization. We can capture this idea by calculating the mutual information (MI) between domains and experts using the probability model described below:

\begin{equation}\label{mi_eq1}
I(D;E) = \Sigma_{i=1}^{M} \Sigma_{j=1}^{N} P(D_i, E_j) log\frac{P(D_i, E_j)}{P(D_i)P(E_j)} .
\end{equation}

If \(D\) and \(E\) are independent random variables, \(p(D, E) = p(D) p(E)\) and then \(I(D; E) = 0\). On the other hand, if each expert is exclusively assigned to a single domain (when M = N), the dependence between the domain and experts (and thus the mutual information) is maximized.
 
In our implementation, a learnable parameter matrix $\mathbf{J} \in \mathbb{R}^{D_M \times D_N}$ denotes the joint-probability matrix $p(D, E)$, and it updates in the dynamic routing matrix in each domain for inputs $X$ for each batch.  Besides, $ \mathbf{d}_k$  and $ \mathbf{g}_k$  denote the domain one-hot index matrix and sparse gating logits matrix in batch $k$, respectively. The updating process of the joint-probability matrix in batch ${k}$ is described as follows:

\begin{equation}\label{eq: mi_j}
\begin{aligned}
    \mathbf{J}_{k} = \beta * \mathbf{J}_{k-1} + (1- \beta) * \mathbf{d}_k \odot \mathbf{g}_k
\end{aligned}
\end{equation}

Finally, the total loss is the combination of $\mathcal{L}^{BCE}$ and mutual information loss \(  \mathcal{L}^{MI}\):

\begin{equation}\label{mi_eq2}
\begin{aligned}
    \mathcal{L}^{MI} = - \sum_{i=1}^{n} \sum_{j=1}^{m}  (\mathbf{J}_{ij}  \cdot log(\frac{\mathbf{J}_{ij}}{\mathbf{P(E)} \cdot \mathbf{P(D)}} )) .
\end{aligned}
\end{equation}





\[ \mathcal{L}^{total} = \mathcal{L}^{BCE} +  \alpha\mathcal{L}^{MI} , \] 

where $\alpha$ is a hyper-parameter to adjust the weight of the mutual information loss $ \mathcal{L}^{MI}$.

%% file: 3.experiments.tex
\section{EXPERIMENTS}
In this section, we conduct empirical studies to demonstrate the effectiveness of the proposed CESAA. We aim to answer the following research questions:
\begin{itemize}
    \item \textbf{RQ1:} How does CESAA perform, comparing with the state-of-the-art multi-domain recommendation methods and other baselines?
    \item \textbf{RQ2:} What is the contribution of key components to the model performance?
\end{itemize}

\subsection{Experimental Settings}
\subsubsection{Dataset Description} In order to verify the effectiveness of the proposed CESAA, we conduct experiments on both retrieval stage and ranking stage datasets. 
\begin{itemize}
    \item \textbf{Industrial dataset}. A retrieval stage dataset. We collect 1.5 billion samples from the OPPO online advertising system that involve 33 domains between 20231201 and 20231218. We apportion the samples into training and testing sets along the time sequence. We evaluate the testing dataset in four groups: all; positive vs. easy negative (PE); positive vs. medium negative (PM); and positive vs. hard negative (PN). Easy/medium/hard negatives denote the randomly selected negative samples, samples from the output of the pre-rank model but not selected by the rank model, and tail samples from the output of the rank model, respectively. In terms of task difficulty, PH > PM > ALL > PE.

    
    \item \textbf{Ali-CCP}\footnote{https://tianchi.aliyun.com/dataset/408}. A ranking stage dataset. Ali-CCP is a public dataset with training and testing set sizes of over 42.3 million and 43 million, respectively.  Following AdaptDHM\cite{li2022adaptdhm}, we attach more features (domain indicator, user gender, user city) to partition samples, resulting in 33 domains. We randomly split each dataset into training, evaluation, and test sets with a ratio of $6:2:2$.  We use \textit{Group AUC} \cite{sheng2021star, zhou2018din}  (GAUC) as the metric.
\end{itemize}


    
\begin{table*}[t]
\caption{Overall performance comparisons on industrial retrieval dataset, we use Req-GAUC and Recall@10-1 as the metric.  } \label{t1}
\begin{tabular}{@{}lllllllll@{}}
\toprule
\hspace{1em} Model      & \multicolumn{4}{c}{Req-GAUC}      & \multicolumn{4}{c}{Recall@10-1}      \\ \midrule
            & ALL    & PE     & PM     & PH     & ALL    & PE     & PM     & PH     \\ \midrule
\hspace{1em}DSSM        & 0.9264 & 0.9967 & 0.7700 & 0.7329 & 0.5998 & 0.9939 & 0.6689 & 0.6841 \\
\hspace{1em}POSO     &  0.9253 &   0.9965 & 0.7661 & 0.7310 & 0.5984 & 0.9934 & 0.6625 &
0.6824 \\
\hspace{1em}PEPNet     &  0.9263 & 0.9966 & 0.7689 & 0.7330 & 0.5997 & 0.9937 & 0.6690 & 0.6838  \\
\hspace{1em}MVKE     &  0.9254    & 0.9966 & 0.7657 & 0.7313 & 0.5935 & 0.994  & 0.6620  & 0.6821    \\

\hspace{1em}\textbf{CESAA}        & \textbf{0.9277} &  \textbf{0.9969} &  \textbf{0.7749} &  \textbf{0.7345} & \textbf{0.6064} &  \textbf{0.9943} &  \textbf{0.6764} &  \textbf{0.6866} \\
\hspace{1em}CESAA w/o \textbf{AEA} & 0.9270 & 0.9968 & 0.7732 & 0.7323 & 0.6029 & 0.9942 & 0.6735 & 0.6844 \\ \midrule
\hspace{1em}Improvement  & 0.14\% & 0.02\% & 0.64\% & 0.22\% & 1.10\% & 0.04\% & 1.12\% & 0.37\% \\ \bottomrule
\end{tabular}
\end{table*}

\subsubsection{Evaluation Metric} 
We  categorize the evaluation metrics used in the experiment into the following three types:

\begin{itemize}

\item \textbf{GAUC}  focus on CTR prediction capability for each user instead of on all samples. The GAUC is defined as:
    \begin{equation}
        \text{GAUC}=\frac{\sum_{u\in\mathcal{U}}w_u\text{AUC}_u}{\sum_{u\in\mathcal{U}}w_u},
    \end{equation}
    where $w_u$ is the weight for user $u$. Here we set $w_u$ as $1$ for each user.

\item \textbf{Req-GAUC} extends the GAUC metric to the query level, taking into account the varying difficulty of different queries.

\item \textbf{Recall@N-K} is defined as the ratio of the intersection of the top N candidates from the candidate list and the top K items from the target list. This ratio indicates the proportion of relevant items that are successfully retrieved within the specified ranks. In our business settings, N is set to 10 and K is set to 1.
\end{itemize}


\subsubsection{Model Comparision}
Considering the differences in scale and task between the two datasets, we set up different baseline models. 
    
    For the Industrial dataset,  in an online retrieval system, the item tower and user tower require a decoupled design to support the storage of item embeddings in Redis, enabling real-time inference for user requests. Moreover, most classic MoE structures like \textbf{PLE}\cite{tang2020ple} have multiple output heads, increasing online storage consumption. Therefore, we choose some commonly used lightweight two-tower MDR works as comparative baselines: \textbf{DSSM}\cite{huang2013learning},  \textbf{POSO}\cite{dai2021poso}, \textbf{PEPNet}\cite{chang2023pepnet}, and \textbf{MVKE}\cite{xu2022mvke}. (1) \textbf{DSSM}. A two-tower model composed of a user tower and an item tower. We additionally add a SE-Block\cite{hu2018squeeze} after the user tower embedding layer. Due to the domain 
    (2) \textbf{POSO}. We add GateNU into the user tower with domain features as GateNU input.  (3) \textbf{PEPNet}. We add GateNU after the embedding layer with domain features and top30\% important features under stop-gradient settings.
    (4) \textbf{MVKE}. After the virtual kernel attention operation in MVKE bottom layers, we design four experts and add a residual connection. 
 
    For ALI-CCP dataset, We compare our approach with the following baselines: (1) \textbf{DNN},  (2) \textbf{MMOE}\cite{ma2018mmoe}, and (3) \textbf{PLE}\cite{tang2020ple} since they are commonly used in MDR. 
\subsubsection{implementation Details} Each MLP in all models has the same depth of hidden layers with 3 layers (256-128-64). All the activation function is set to ReLU. We apply the Adam \cite{kingma2014adam}  as optimizer with a batch size of 1024. The learning rate is set to 1e-3.  
The default numbers of sparse experts and TopK of CESAA are set to 4 and 3, respectively. 

\subsection{Experimental Results}
\begin{figure}
  \centering
  \includegraphics[width=0.9\columnwidth]{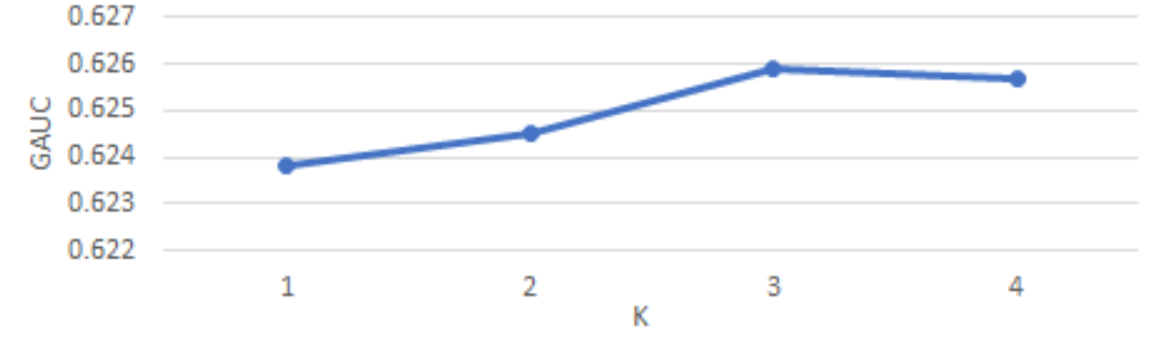}
  \caption{Comparison of different selected sparse expert number $K$ of CESAA in Ali-CCP dataset.}
  \label{fig:sparse_expert_exp}
\end{figure}

\subsubsection{Effectiveness Verification}

 Table \ref{t1} summarizes the experimental result on the retrieval stage dataset. 
 From an overall perspective (ALL), CESAA achieves best performance among all methods with improvements over 0.1\% and 1\% on Req-GAUC and Recall@N-K metrics respectively.
 Additionally, CESAA is comprehensively superior to other baselines on fine-grained level (PE/PM/PH). This indicate that regardless of the difficulty of the given negative samples, CESAA is capable of distinguishing positive samples from them more effectively.
 
 Furthermore, CESAA is superior to dynamic weight-based methods, such as POSO and EPNet, indicating that they fail to model diversity when faced with such large numbers of domains. Similarly, compared with the deployed DSSM, MVKE decreased by 0.10\% and 0.63\% in Req-GAUC and Recall@N-K respectively. 
 This indicates that relying solely on the implicit learning of the gating network makes it difficult to obtain the optimal selection strategy between different scenarios and experts, inducing limited performance.

 Table \ref{t2} shows the results of CESAA on the ranking stage experiment, which clearly proves the efficiency of CESAA above other models. CESAA achieves up to 0.11\% improvement over MMOE and 0.25\% over PLE .
\subsubsection{Ablation Study} We set up four ablation experiments to verify the key components of CESAA. \textbf{MOEwa} denotes MMOE with AEA module. \textbf{CESAAa} and \textbf{CESAAs} denote that we drop out the AEA module and the shared expert in CES module, respectively. \textbf{CESAAas} denotes that both of them are dropped out simultaneously.  Table \ref{t2} also shows the importance of mutual information constraint loss and share experts. Mutual information constraint loss plays an important role in the proposed method, with 0.08\% improvement over MMOE and 0.06\% improvement over CESAA. While the shared expert design also improves CESAA with 0.05\% improvements. Moreover, Fig \ref{fig:sparse_expert_exp} shows that as the number of selected sparse experts increases, the performance of CESAA increases overall, and activating all experts may not be beneficial for every individual instance.






\begin{table}[!t]
\scriptsize
\caption{Performance comparison on Ali-CCP dataset.  } \label{t2}
\begin{tabular*}{\linewidth}{@{\extracolsep{\fill}}ccccccccc@{}}
\toprule
Model & DNN  & PLE & MMOE & CESAA &  MOEwa & CESAAa & CESAAs  & CESAAas \\ \midrule
GAUC   & 0.6222  & 0.6232 & 0.6246 & \textbf{0.6257} & 0.6254  & 0.6253  & 0.6254 & 0.6238  \\ 

\end{tabular*}
\end{table}







%% file: main.bbl

\begin{thebibliography}{15}


\ifx \showCODEN    \undefined \def \showCODEN     #1{\unskip}     \fi
\ifx \showDOI      \undefined \def \showDOI       #1{#1}\fi
\ifx \showISBNx    \undefined \def \showISBNx     #1{\unskip}     \fi
\ifx \showISBNxiii \undefined \def \showISBNxiii  #1{\unskip}     \fi
\ifx \showISSN     \undefined \def \showISSN      #1{\unskip}     \fi
\ifx \showLCCN     \undefined \def \showLCCN      #1{\unskip}     \fi
\ifx \shownote     \undefined \def \shownote      #1{#1}          \fi
\ifx \showarticletitle \undefined \def \showarticletitle #1{#1}   \fi
\ifx \showURL      \undefined \def \showURL       {\relax}        \fi
\providecommand\bibfield[2]{#2}
\providecommand\bibinfo[2]{#2}
\providecommand\natexlab[1]{#1}
\providecommand\showeprint[2][]{arXiv:#2}

\bibitem[Chang et~al\mbox{.}(2023)]%
        {chang2023pepnet}
\bibfield{author}{\bibinfo{person}{Jianxin Chang}, \bibinfo{person}{Chenbin Zhang}, \bibinfo{person}{Yiqun Hui}, \bibinfo{person}{Dewei Leng}, \bibinfo{person}{Yanan Niu}, \bibinfo{person}{Yang Song}, {and} \bibinfo{person}{Kun Gai}.} \bibinfo{year}{2023}\natexlab{}.
\newblock \showarticletitle{Pepnet: Parameter and embedding personalized network for infusing with personalized prior information}. In \bibinfo{booktitle}{\emph{Proceedings of the 29th ACM SIGKDD Conference on Knowledge Discovery and Data Mining}}. \bibinfo{pages}{3795--3804}.
\newblock


\bibitem[Chen et~al\mbox{.}(2023)]%
        {chen2023mod}
\bibfield{author}{\bibinfo{person}{Zitian Chen}, \bibinfo{person}{Yikang Shen}, \bibinfo{person}{Mingyu Ding}, \bibinfo{person}{Zhenfang Chen}, \bibinfo{person}{Hengshuang Zhao}, \bibinfo{person}{Erik~G Learned-Miller}, {and} \bibinfo{person}{Chuang Gan}.} \bibinfo{year}{2023}\natexlab{}.
\newblock \showarticletitle{Mod-squad: Designing mixtures of experts as modular multi-task learners}. In \bibinfo{booktitle}{\emph{Proceedings of the IEEE/CVF Conference on Computer Vision and Pattern Recognition}}. \bibinfo{pages}{11828--11837}.
\newblock


\bibitem[Dai et~al\mbox{.}(2021)]%
        {dai2021poso}
\bibfield{author}{\bibinfo{person}{Shangfeng Dai}, \bibinfo{person}{Haobin Lin}, \bibinfo{person}{Zhichen Zhao}, \bibinfo{person}{Jianying Lin}, \bibinfo{person}{Honghuan Wu}, \bibinfo{person}{Zhe Wang}, \bibinfo{person}{Sen Yang}, {and} \bibinfo{person}{Ji Liu}.} \bibinfo{year}{2021}\natexlab{}.
\newblock \showarticletitle{POSO: personalized cold start modules for large-scale recommender systems}.
\newblock \bibinfo{journal}{\emph{arXiv preprint arXiv:2108.04690}} (\bibinfo{year}{2021}).
\newblock


\bibitem[Hu et~al\mbox{.}(2018)]%
        {hu2018squeeze}
\bibfield{author}{\bibinfo{person}{Jie Hu}, \bibinfo{person}{Li Shen}, {and} \bibinfo{person}{Gang Sun}.} \bibinfo{year}{2018}\natexlab{}.
\newblock \showarticletitle{Squeeze-and-excitation networks}. In \bibinfo{booktitle}{\emph{Proceedings of the IEEE conference on computer vision and pattern recognition}}. \bibinfo{pages}{7132--7141}.
\newblock


\bibitem[Huang et~al\mbox{.}(2013)]%
        {huang2013learning}
\bibfield{author}{\bibinfo{person}{Po-Sen Huang}, \bibinfo{person}{Xiaodong He}, \bibinfo{person}{Jianfeng Gao}, \bibinfo{person}{Li Deng}, \bibinfo{person}{Alex Acero}, {and} \bibinfo{person}{Larry Heck}.} \bibinfo{year}{2013}\natexlab{}.
\newblock \showarticletitle{Learning deep structured semantic models for web search using clickthrough data}. In \bibinfo{booktitle}{\emph{Proceedings of the 22nd ACM international conference on Information \& Knowledge Management}}. \bibinfo{pages}{2333--2338}.
\newblock


\bibitem[Jiang et~al\mbox{.}(2022)]%
        {jiang2022adin}
\bibfield{author}{\bibinfo{person}{Yuchen Jiang}, \bibinfo{person}{Qi Li}, \bibinfo{person}{Han Zhu}, \bibinfo{person}{Jinbei Yu}, \bibinfo{person}{Jin Li}, \bibinfo{person}{Ziru Xu}, \bibinfo{person}{Huihui Dong}, {and} \bibinfo{person}{Bo Zheng}.} \bibinfo{year}{2022}\natexlab{}.
\newblock \showarticletitle{Adaptive domain interest network for multi-domain recommendation}. In \bibinfo{booktitle}{\emph{Proceedings of the 31st ACM International Conference on Information \& Knowledge Management}}. \bibinfo{pages}{3212--3221}.
\newblock


\bibitem[Kingma and Ba(2014)]%
        {kingma2014adam}
\bibfield{author}{\bibinfo{person}{Diederik~P Kingma} {and} \bibinfo{person}{Jimmy Ba}.} \bibinfo{year}{2014}\natexlab{}.
\newblock \showarticletitle{Adam: A method for stochastic optimization}.
\newblock \bibinfo{journal}{\emph{arXiv preprint arXiv:1412.6980}} (\bibinfo{year}{2014}).
\newblock


\bibitem[Li et~al\mbox{.}(2022)]%
        {li2022adaptdhm}
\bibfield{author}{\bibinfo{person}{Jinyun Li}, \bibinfo{person}{Huiwen Zheng}, \bibinfo{person}{Yuanlin Liu}, \bibinfo{person}{Minfang Lu}, \bibinfo{person}{Lixia Wu}, {and} \bibinfo{person}{Haoyuan Hu}.} \bibinfo{year}{2022}\natexlab{}.
\newblock \showarticletitle{AdaptDHM: Adaptive Distribution Hierarchical Model for Multi-Domain CTR Prediction}.
\newblock \bibinfo{journal}{\emph{arXiv preprint arXiv:2211.12105}} (\bibinfo{year}{2022}).
\newblock


\bibitem[Ma et~al\mbox{.}(2018)]%
        {ma2018mmoe}
\bibfield{author}{\bibinfo{person}{Jiaqi Ma}, \bibinfo{person}{Zhe Zhao}, \bibinfo{person}{Xinyang Yi}, \bibinfo{person}{Jilin Chen}, \bibinfo{person}{Lichan Hong}, {and} \bibinfo{person}{Ed~H Chi}.} \bibinfo{year}{2018}\natexlab{}.
\newblock \showarticletitle{Modeling task relationships in multi-task learning with multi-gate mixture-of-experts}. In \bibinfo{booktitle}{\emph{Proceedings of the 24th ACM SIGKDD international conference on knowledge discovery \& data mining}}. \bibinfo{pages}{1930--1939}.
\newblock


\bibitem[Shazeer et~al\mbox{.}(2017)]%
        {shazeer2017outrageously}
\bibfield{author}{\bibinfo{person}{Noam Shazeer}, \bibinfo{person}{Azalia Mirhoseini}, \bibinfo{person}{Krzysztof Maziarz}, \bibinfo{person}{Andy Davis}, \bibinfo{person}{Quoc Le}, \bibinfo{person}{Geoffrey Hinton}, {and} \bibinfo{person}{Jeff Dean}.} \bibinfo{year}{2017}\natexlab{}.
\newblock \showarticletitle{Outrageously large neural networks: The sparsely-gated mixture-of-experts layer}.
\newblock \bibinfo{journal}{\emph{arXiv preprint arXiv:1701.06538}} (\bibinfo{year}{2017}).
\newblock


\bibitem[Sheng et~al\mbox{.}(2021)]%
        {sheng2021star}
\bibfield{author}{\bibinfo{person}{Xiang-Rong Sheng}, \bibinfo{person}{Liqin Zhao}, \bibinfo{person}{Guorui Zhou}, \bibinfo{person}{Xinyao Ding}, \bibinfo{person}{Binding Dai}, \bibinfo{person}{Qiang Luo}, \bibinfo{person}{Siran Yang}, \bibinfo{person}{Jingshan Lv}, \bibinfo{person}{Chi Zhang}, \bibinfo{person}{Hongbo Deng}, {et~al\mbox{.}}} \bibinfo{year}{2021}\natexlab{}.
\newblock \showarticletitle{One model to serve all: Star topology adaptive recommender for multi-domain ctr prediction}. In \bibinfo{booktitle}{\emph{Proceedings of the 30th ACM International Conference on Information \& Knowledge Management}}. \bibinfo{pages}{4104--4113}.
\newblock


\bibitem[Tang et~al\mbox{.}(2020)]%
        {tang2020ple}
\bibfield{author}{\bibinfo{person}{Hongyan Tang}, \bibinfo{person}{Junning Liu}, \bibinfo{person}{Ming Zhao}, {and} \bibinfo{person}{Xudong Gong}.} \bibinfo{year}{2020}\natexlab{}.
\newblock \showarticletitle{Progressive layered extraction (ple): A novel multi-task learning (mtl) model for personalized recommendations}. In \bibinfo{booktitle}{\emph{Proceedings of the 14th ACM Conference on Recommender Systems}}. \bibinfo{pages}{269--278}.
\newblock


\bibitem[Xu et~al\mbox{.}(2022)]%
        {xu2022mvke}
\bibfield{author}{\bibinfo{person}{Zhenhui Xu}, \bibinfo{person}{Meng Zhao}, \bibinfo{person}{Liqun Liu}, \bibinfo{person}{Lei Xiao}, \bibinfo{person}{Xiaopeng Zhang}, {and} \bibinfo{person}{Bifeng Zhang}.} \bibinfo{year}{2022}\natexlab{}.
\newblock \showarticletitle{Mixture of virtual-kernel experts for multi-objective user profile modeling}. In \bibinfo{booktitle}{\emph{Proceedings of the 28th ACM SIGKDD Conference on Knowledge Discovery and Data Mining}}. \bibinfo{pages}{4257--4267}.
\newblock


\bibitem[Zhou et~al\mbox{.}(2018)]%
        {zhou2018din}
\bibfield{author}{\bibinfo{person}{Guorui Zhou}, \bibinfo{person}{Xiaoqiang Zhu}, \bibinfo{person}{Chenru Song}, \bibinfo{person}{Ying Fan}, \bibinfo{person}{Han Zhu}, \bibinfo{person}{Xiao Ma}, \bibinfo{person}{Yanghui Yan}, \bibinfo{person}{Junqi Jin}, \bibinfo{person}{Han Li}, {and} \bibinfo{person}{Kun Gai}.} \bibinfo{year}{2018}\natexlab{}.
\newblock \showarticletitle{Deep interest network for click-through rate prediction}. In \bibinfo{booktitle}{\emph{Proceedings of the 24th ACM SIGKDD international conference on knowledge discovery \& data mining}}. \bibinfo{pages}{1059--1068}.
\newblock


\bibitem[Zhou et~al\mbox{.}(2022)]%
        {zhou2022mixture}
\bibfield{author}{\bibinfo{person}{Yanqi Zhou}, \bibinfo{person}{Tao Lei}, \bibinfo{person}{Hanxiao Liu}, \bibinfo{person}{Nan Du}, \bibinfo{person}{Yanping Huang}, \bibinfo{person}{Vincent Zhao}, \bibinfo{person}{Andrew~M Dai}, \bibinfo{person}{Quoc~V Le}, \bibinfo{person}{James Laudon}, {et~al\mbox{.}}} \bibinfo{year}{2022}\natexlab{}.
\newblock \showarticletitle{Mixture-of-experts with expert choice routing}.
\newblock \bibinfo{journal}{\emph{Advances in Neural Information Processing Systems}}  \bibinfo{volume}{35} (\bibinfo{year}{2022}), \bibinfo{pages}{7103--7114}.
\newblock


\end{thebibliography}
